%% file: adaie_preprint.tex
\documentclass{article} %
\pdfoutput=1
\usepackage{iclr2027_conference,times}
\usepackage{algorithm}
\usepackage{algpseudocode}
\usepackage{graphicx}
\usepackage{wrapfig}
\usepackage{booktabs}
\usepackage[table]{xcolor}
\usepackage{multirow} 
\usepackage{amsthm}
\usepackage{capt-of}

\newcommand{\mat}[1]{\mathbf{#1}}

\input{math_commands.tex}

\usepackage{hyperref}
\usepackage{url}

\title{Controlling Action-Head Co-Adaptation to VLM Representation Anisotropy}

\author{%
Ruijie Zhang$^{1}$\thanks{Work done during an internship at XPeng U.S.}, Jiayin Zheng$^{2}$, Yequan Zhao$^{1}$, Liyan Tan$^{1}$, \\
\bf Ziyue Liu$^{1}$, Zhengyang Wang$^{1}$, Kunpeng Song$^{2}$, Nathan Zhao$^{2}$, \\
\bf Jian Yao$^{2}$, Yin Zhang$^{2}$, Zheng Zhang$^{1}$ \\[0.6ex]
{\normalfont $^{1}$University of California, Santa Barbara \qquad $^{2}$Foundation Model Team, XPeng} \\
{\normalfont\texttt{ruijiezhang@ucsb.edu}}
}

\iclrfinalcopy
\begin{document}

\maketitle
\lhead{Preprint. Under review.}
\begin{abstract}
Modern learning systems increasingly combine pretrained models with downstream heads. Despite the growing importance of such modular designs, how the downstream model co-adapts to the geometry of pretrained representations remains relatively underexplored. We study this problem through vision-language-action models (VLAs), where a pretrained vision-language model (VLM) conditions an action model for control. We find that the action model progressively concentrates its sensitivity on high-variance dimensions of the VLM outputs. Through controlled experiments across different backbone-head models, we further show that counteracting this sensitivity concentration can be beneficial under representation--task mismatch settings. Motivated by these observations, we introduce \emph{Adaptive Interface Equalizer} (AdaIE), a parameter-free mechanism that reweights pretrained representations according to the evolving downstream sensitivity during training. AdaIE improves VLA success rates on simulation benchmarks, with gains of up to  $+5.28$ percentage points (pp) on RoboCasa-GR1 and $+1.68$ pp on RoboTwin 2.0, and achieves an average gain of $+19$ pp across five real-world manipulation tasks. It further generalizes to diffusion policies and world models, improving downstream performance in both settings. Our results highlight interface co-adaptation as an important optimization problem when coupling pretrained representations with downstream models.
\end{abstract}

\section{Introduction}
VLAs transfer pretrained VLM knowledge to robotic control~\citep{ma2026survey,kim2024openvla}. Recent approaches use diffusion- or flow-matching-based action heads to map VLM representations of observations and instructions to continuous actions~\citep{black2024pi_0,li2024cogact,chi2025diffusion,bjorck2025gr00t}. Despite substantial progress in data pipelines, architectures, and training~\citep{brohan2023rt,intelligence2025pi_,hou2025dita,li2026matters}, the representation interface connecting VLMs and action heads remains comparatively underexplored.

The VLM and action model are typically connected through a lightweight conditioning interface, such as cross-attention and concatenation. Through this interface, representations produced by the VLM are directly consumed by the action model for action learning. Despite its simplicity, this interface determines how the representation geometry learned from vision-language pretraining is exposed to the action head. 

In this work, begin by studying this interface in VLAs, we find that VLM outputs exhibit strong channel-wise anisotropy, and that the action model progressively becomes more sensitive to these dimensions during training (See Figure ~\ref{fig:energy_sensitivity_alignment}). However, the two components are optimized for different objectives: the VLM is pretrained to organize information for vision-language understanding, whereas the action head is trained for control. We therefore ask:

\textbf{\textit{Why should an action head trained for control {increasingly} rely on the same feature dimensions shaped by a VLM pretrained for a different objective?}}

To answer this question, we study the interface by tracking VLM channel variance and action-head sensitivity throughout training, and further compare sensitivity reweighting across interfaces with different degrees of representation--task alignment. We find that counteracting the evolving sensitivity imbalance consistently improves performance under representation--task mismatch.

Motivated by these observations, we introduce \emph{Adaptive Interface Equalizer} (AdaIE), a parameter-free mechanism that reweights representations using evolving downstream sensitivity. We summarize our contributions as follows:
\begin{itemize}
    \item We identify backbone--head co-adaptation: the head progressively concentrates its sensitivity on high-variance dimensions of pretrained backbone representations. We show that reducing sensitivity concentration can improve learning under representation--task mismatch.
    
    \item We propose \emph{Adaptive Interface Equalizer} (AdaIE), a parameter-free mechanism that adaptively reweights backbone representations according to the evolving head sensitivity to regulate interface co-adaptation.
    
    \item We demonstrate AdaIE's effectiveness in VLAs, with success-rate gains of up to +5.28 percentage points (pp) on RoboCasa-GR1, +1.68 pp on RoboTwin 2.0, and an average of +19 pp across real-world manipulation tasks. We further show that its benefits extend to diffusion policies and world models.
\end{itemize}

\section{Action-Head Co-Adaptation to VLM Anisotropy}

We begin by examining the VLM--DiT interface, characterizing VLM channel-wise anisotropy and the evolution of action-head sensitivity throughout training in Section~\ref{sec:vlm_anisotropic_coadaptation}. We then investigate whether this emerging co-adaptation is beneficial for downstreaming learning in Section~\ref{sec:probe}.

\subsection{VLM Representation Anisotropy and Action-Head Co-Adaptation}
\label{sec:vlm_anisotropic_coadaptation}

\begin{figure}[h]
    \centering
    \includegraphics[width=\linewidth]{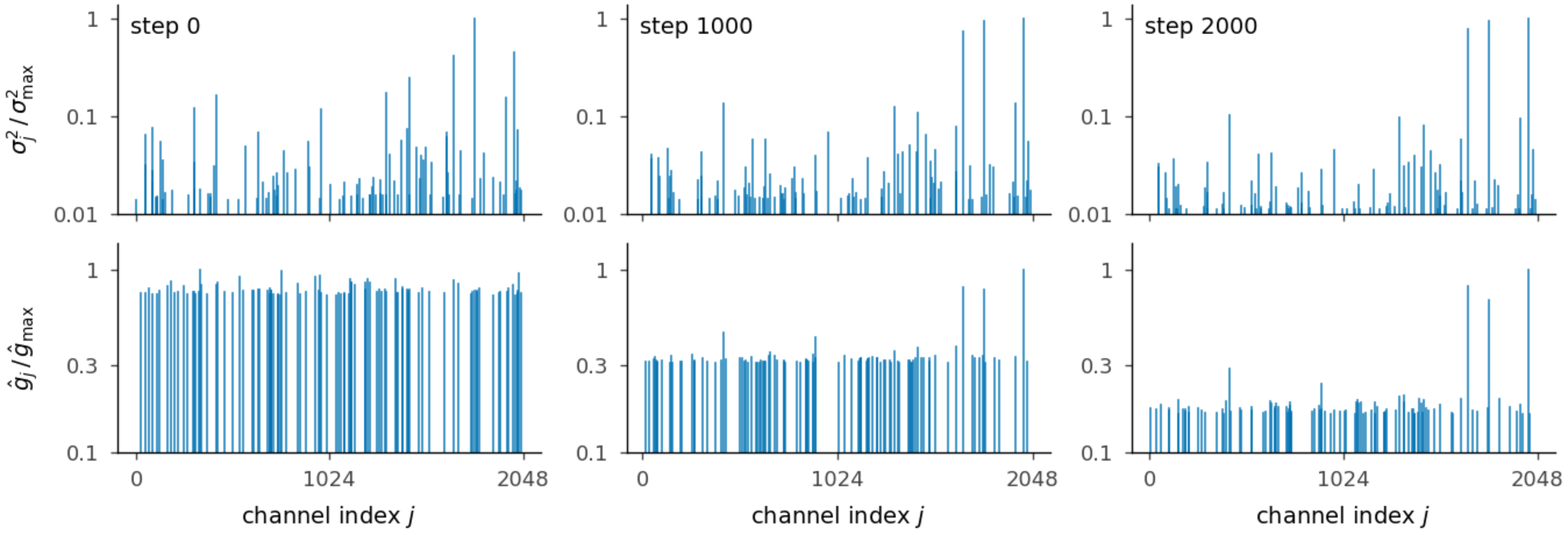}
    \caption{
    Action-head sensitivity progressively aligns with the anisotropic VLM
    representation during training.
    \textbf{Top}: per-channel variance of the VLM output dimensions
    at different training steps.
    \textbf{Bottom}: corresponding per-channel action-head sensitivity.
    While sensitivity is relatively uniform early in training, it progressively
    concentrates on high-variance VLM dimensions as training proceeds, revealing
    increasing alignment between VLM anisotropy and action-head sensitivity.
    }
    \label{fig:energy_sensitivity_alignment}
\end{figure}

We first introduce the notation used throughout this section. Let $\mat{C}\in\mathbb{R}^{N\times D}$ denote the VLM conditioning tokens, where $N$ is the number of tokens and $D$ is the hidden dimension. We use $i\in\{1,\ldots,N\}$ to index conditioning tokens and $j\in\{1,\ldots,D\}$ to index feature channels, such that $\mat{C}_{i,j}$ denotes the activation of channel $j$ at token $i$. Let $\mat{X}\in\mathbb{R}^{N\times D}$ denote input tokens of the action DiT. In the standard VLAs, $\mat{X}=\mat{C}$.

With this notation, we track two per-channel statistics throughout training:
the variance of the VLM representation and the sensitivity of the action head
to each conditioning channel.

\textbf{VLM anisotropy.}
We characterize the anisotropy of VLM representations via
per-channel variance,
\[
\sigma_j^2 = \mathrm{Var}[\mat{C}_{:,j}].
\]
The variance is estimated over conditioning tokens. A highly uneven variance across channels indicates an anisotropic representation, where a small subset of dimensions varies more than the rest.

\textbf{Head sensitivity.}
We characterize the sensitivity of the head model (e.g. DiT action head) to feature channel $j$ of its input $\mat{X}$ by averaging its sensitivity across conditioning tokens. For token $i$, let
\[
\mat{J}_i = \frac{\partial \mat{\hat{a}}}{\partial \mat{X}_{i,:}}
\]
denote the Jacobian of the action prediction with respect to its $D$ feature
channels. We define the sensitivity of channel $j$ as
\[
\mat{g}_j =
\mathbb{E}_{i}\!\left[
\mathrm{diag}\!\left(\mat{J}_i^\top \mat{J}_i\right)_j
\right],
\]
where the expectation averages over conditioning tokens $i$. To estimate $\mat{g_j}$ efficiently, we use Hutchinson-style stochastic probing~\citep{hutchinson1989stochastic}(Algorithm~\ref{alg:sensitivity}).

To study those two metrics, we adopt a VLM--DiT architecture following GR00T-N1~\citep{bjorck2025gr00t}, where vision-language representations condition a flow-matching DiT action head through cross-attention. Following the RoboCasa-GR1 training setup in StarVLA~\citep{community2026starvla}, we use a pretrained Qwen3-VL~\citep{bai2025qwen3} as the VLM backbone and train the action DiT from scratch. For clarity, Figure~\ref{fig:energy_sensitivity_alignment} visualizes the top-100 VLM channels with the highest variance.

As shown in Figure~\ref{fig:energy_sensitivity_alignment}, the VLM exhibits a strongly anisotropic channel-wise variance, while the action-head sensitivity is relatively uniform early in training. As the training goes, the action head becomes increasingly sensitive to the same high-variance VLM dimensions. This growing alignment suggests that the action head progressively co-adapts to the anisotropic structure of the pretrained VLM representation. 
A similar phenomenon is also observed in other models (see Appendix~\ref{app:co-adaption}).

\subsection{Is Co-Adaptation Beneficial for Action Learning?}
\label{sec:probe}

The emerging co-adaptation has two opposing interpretations. One is that the action head becomes more sensitive to high-variance VLM dimensions because these dimensions are more useful for downstream control. An alternative view is that the action head is increasingly shaped by the anisotropic geometry inherited from vision-language pretraining, which may not be optimal for action learning because the VLM and action model are optimized for different domains and objectives. Under this view, reducing the emerging sensitivity imbalance may instead be beneficial.

To distinguish between these two interpretations, we construct a controlled experiment on the sensitivity. We compare \emph{Exploit}, which reinforces the emerging sensitivity pattern, with \emph{Equalize}, which reduces it, using the original training as the \emph{Baseline}.

Specifically, given the estimated sensitivity $\mat{\hat g_j}$, we assign each channel a sensitivity-dependent weight
\begin{equation}
    \mat{s}_j = (\mat{\hat g}_j)^{-\alpha},
    \label{eq:eq_exploit}
\end{equation}
where $\alpha>0$ gives \emph{Equalize}, assigning relatively smaller weights to more-sensitive channels, while $\alpha<0$ gives \emph{Exploit}, assigning relatively larger weights to them.

To examine when such co-adaptation is beneficial or harmful, we consider two types of backbone--head interfaces with different degrees of representation--task alignment:

\begin{itemize}
    \item \textbf{Aligned.}
    The pretrained representation is expected to be well suited to the downstream task. We consider (1) DINOv2-S with a DPT head~\citep{yang2024depth} on NYUv2~\citep{silberman2012indoor} for depth estimation, and (2) DeiT~\citep{touvron2021training} on CIFAR-100~\citep{Krizhevsky09learningmultiple} for image classification.

    \item \textbf{Mismatched.}
    The pretrained representation is learned for an objective substantially different from downstream action generation. 
    We consider
    (1) Diffusion Policy~\citep{chi2025diffusion} on Push-T for visuomotor action learning, and
    (2) GR00T~\citep{bjorck2025gr00t} on RoboCasa-GR1~\citep{robocasa2024} for vision-language-action learning.
\end{itemize}

To study how downstream heads co-adapt to pretrained representations, we initialize the backbone in each of the four tasks with pretrained weights and otherwise follow the official training recipe. 
Within each task, all compared variants use the same configuration, differing only in the value of $\alpha$. Detailed training setups are provided in Appendix~\ref{app:prob_detals}.

The results are shown in Table~\ref{tab:aligned_vs_mismatch}. For the aligned tasks, sensitivity
reweighting has only a marginal effect on performance, with all three variants performing similarly. While in the mismatched tasks exhibit a clear and consistent trend: \emph{Equalize} outperforms the original \emph{Baseline}, while \emph{Exploit} degrades performance. This contrast indicates that co-adaptation to the pretrained representation may be more problematic when the representation is less aligned with the downstream objective.

\begin{table}[t]
\caption{
Effect of sensitivity reweighting under aligned and mismatched interfaces.
}
\centering
\small
\setlength{\tabcolsep}{3.5pt}

\begin{tabular}{lllcccc}
\toprule
\textbf{Task}
& \textbf{Dataset}
& \textbf{Interface}
& \textbf{Metric}
& \textbf{Equalize}
& \textbf{Baseline}
& \textbf{Exploit} \\
\midrule

\multicolumn{7}{l}{\textbf{Aligned}} \\
\addlinespace[2pt]

Depth
& NYUv2
& DINOv2-S $\rightarrow$ DPT
& $\delta_1$ (\%) $\uparrow$
& 93.4\ensuremath{\pm} 0.3
& 93.5\ensuremath{\pm} 0.1
& 93.2\ensuremath{\pm} 0.3 \\

Classification
& CIFAR-100
& DeiT-S $\rightarrow$ Cls.
& Acc. (\%)$\uparrow$
& 90.0 \ensuremath{\pm} 0.2
& 89.6 \ensuremath{\pm} 0.4
& 89.8 \ensuremath{\pm} 0.4 \\

\midrule

\multicolumn{7}{l}{\textbf{Mismatched}} \\
\addlinespace[2pt]
Diffusion Policy
& Push-T
& ResNet $\rightarrow$ DP
& Score\ensuremath{\uparrow}
& 0.936 \ensuremath{\pm} 0.003
& 0.922 \ensuremath{\pm} 0.003
& 0.909 \ensuremath{\pm} 0.009 \\

VLA
& RoboCasa
& VLM $\rightarrow$ DiT
& Succ. (\%)$\uparrow$
& 22.9 \ensuremath{\pm} 0.5
& 19.4 \ensuremath{\pm} 0.6
& 16.2 \ensuremath{\pm} 1.1\\

\bottomrule
\end{tabular}

\label{tab:aligned_vs_mismatch}
\end{table}

Motivated by the observations above, we propose \emph{Adaptive Interface Equalizer} (AdaIE) to improve backbone--head training under representation--task mismatch.

\section{Adaptive Interface Equalizer}
\label{sec:adaie}
We now introduce \emph{Adaptive Interface Equalizer} (AdaIE), a parameter-free mechanism that continuously compensates for the evolving sensitivity imbalance during training. AdaIE consists of two steps:sensitivity estimation (Section~\ref{sec:sensitivity_estimation}) and sensitivity equalization (Section~\ref{sec:sensitivity_eq}). We adopt VLA as the representive model to introduce AdaIE.
\begin{figure}[h]
    \centering
    \includegraphics[width=\linewidth]{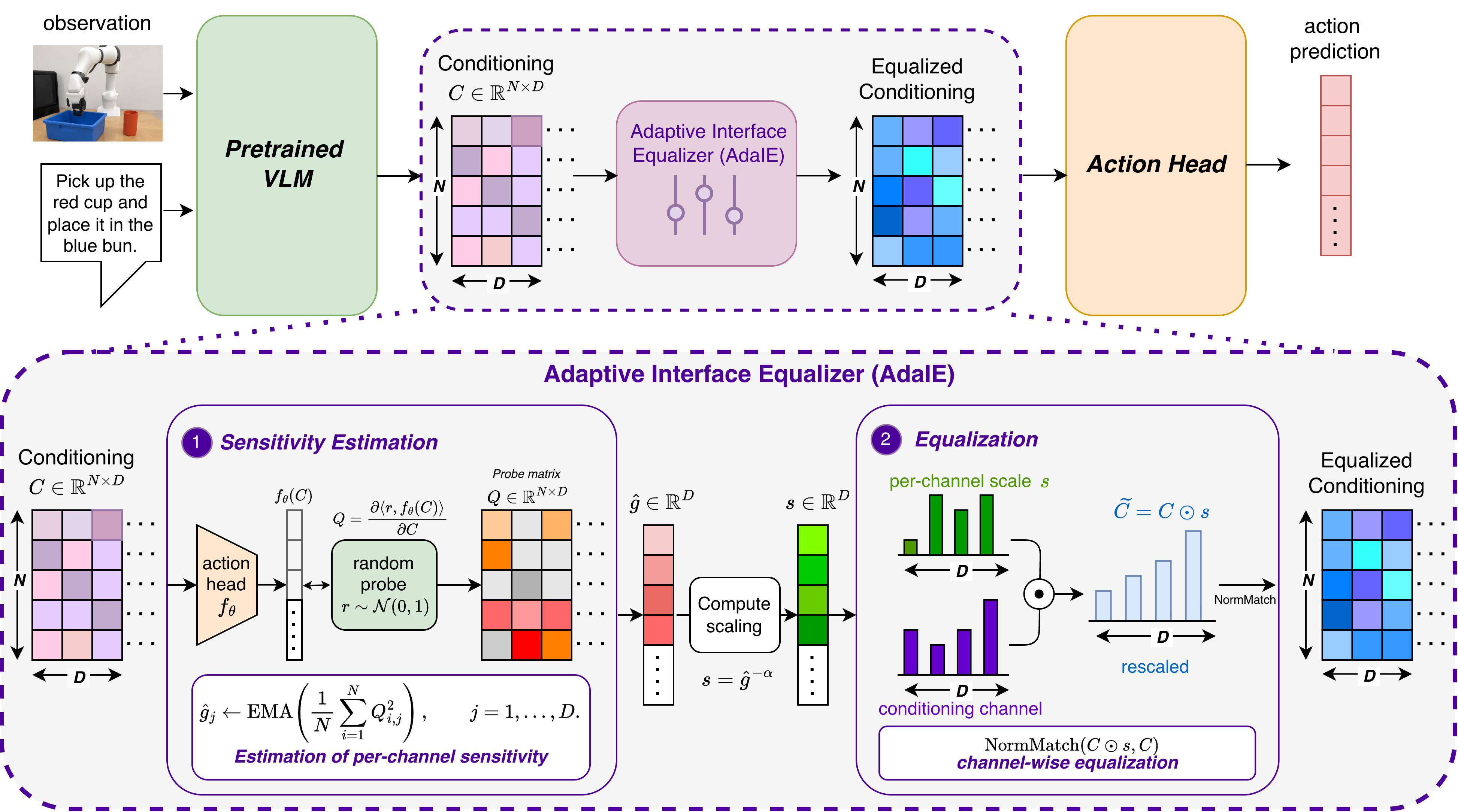}
    \caption{
Overview of \textsc{Adaptive Interface Equalizer} (AdaIE) on VLAs.
AdaIE operates at the interface between the VLM and the action DiT.
It estimates the action head's per-channel sensitivity online using stochastic probes, maintains a running sensitivity estimate, and inversely
rescales the VLM conditioning channels before they are consumed by the DiT.
}
    \label{fig:main}
\end{figure}
\subsection{Sensitivity Estimation}
\label{sec:sensitivity_estimation}

Let $\mat{C}\in\mathbb{R}^{N\times D}$ denote the VLM conditioning at the VLM--DiT interface. As defined in Section~\ref{sec:vlm_anisotropic_coadaptation}, we measure the DiT sensitivity of channel $j$ by the squared Jacobian norm of the action prediction with respect to that channel. Computing the full Jacobian is prohibitively expensive, so we estimate its diagonal online using Hutchinson's random probe~\citep{hutchinson1989stochastic}. Specifically, for a random output-space direction $\mat{r}\sim\mathcal{N}(0,I)$ and a action head $f_\theta$, we compute the
\textit{sensitivity probe matrix}
\begin{equation}
    \mat{Q} =
    \frac{\partial \langle \mat{r}, f_\theta(\mat{C})\rangle}{\partial \mat{C}}
    \in \mathbb{R}^{N\times D}.
\end{equation}
The per-channel sensitivity estimate is
\begin{equation}
    \mat{g}_j =
    \frac{1}{N}\sum_{i=1}^{N} \mat{Q}_{i,j}^2,
\end{equation}
which is then exponentially averaged during training to obtain $\hat g_j$.
Algorithm~\ref{alg:sensitivity} summarizes the full process. In practice, the
probe is performed every $K$ optimization steps rather than at every step,
introducing little additional training overhead.

\begin{algorithm}[h]
\caption{Sensitivity Estimation}
\label{alg:sensitivity}
\begin{algorithmic}[1]
\Require VLM conditioning $\mat{C}\in\mathbb{R}^{N\times D}$;
         action head $f_\theta$; EMA sensitivity $\mat{\hat g}\in\mathbb{R}^{D}$
\Ensure Updated sensitivity $\mat{\hat g}$

\State $r \sim \mathcal{N}(0,I)$
\Comment{random output-space probe}

\State $\mat{Q} \leftarrow
\frac{\partial \langle r, f_\theta(\mat{C})\rangle}{\partial \mat{C}}$
\Comment{sensitivity probe matrix}

\State $\mat{g} \leftarrow
\mathrm{mean}_{i=1}^{N}\!\left(\mat{Q}_i^2\right)$
\Comment{per-channel sensitivity}

\State $\mat{\hat g} \leftarrow
\rho\,\mat{\hat g} + (1-\rho)\,\mat{g}$
\Comment{EMA over training}

\State \Return $\mat{\hat g}$
\end{algorithmic}
\end{algorithm}

\subsection{Sensitivity Equalization}
\label{sec:sensitivity_eq}

Given the estimated sensitivity $\mat{\hat g}_j$, AdaIE assigns each conditioning
channel a sensitivity-dependent scaling factor:
\begin{equation}
    \mat{s}_j = (\mat{\hat g}_j)^{-\alpha},
    \qquad
    \mat{\widetilde C}_{i,j}=\mat{C}_{i,j}\mat{s}_j,
    \label{eq:adaie}
\end{equation}
where $\alpha\geq0$ controls the equalization strength. Larger $\alpha$ produces stronger equalization, while $\alpha=0$ recovers the original conditioning.

Channel-wise rescaling can change the overall magnitude of the conditioning.
We therefore match the magnitude of the transformed representation to that of
the original VLM output:
\begin{equation}
    \mat{Z} = \operatorname{NormMatch}( \mat{\widetilde C},\mat{C}).
\end{equation}
This removes the global scale change introduced by Eq.~\ref{eq:adaie}, leaving only the relative reweighting across channels. In AdaIE, we apply Algorithm~\ref{alg:sensitivity} to the reweighted and norm-matched conditioning $\mat{Z}$, and the scaling factors for each forward pass are computed from the EMA of preceding sensitivity probes. We summarize AdaIE in Figure~\ref{fig:main} and Algorithm~\ref{alg:adaie}. The transformation introduces no learnable parameters, and its sensitivity statistics are updated online jointly with VLA training, and frozen during inference.

\begin{algorithm}[H]
\caption{\textsc{Adaptive Interface Equalizer} (AdaIE)}
\label{alg:adaie}
\begin{algorithmic}[1]
\Require VLM output $\mat{C}\in\mathbb{R}^{N\times D}$;
         sensitivity $\mat{\hat g}\in\mathbb{R}^{D}$;
         equalization strength $\alpha$;
         stabilization constant $\epsilon$
\Ensure DiT conditioning $\mat{Z}\in\mathbb{R}^{N\times D}$

\State $s \leftarrow (\hat g + \epsilon)^{-\alpha}$
\Comment{sensitivity-based scaling}

\State $\mat{\widetilde C} \leftarrow \mat{C}\odot s$
\Comment{channel-wise reweighting}

\State $\mat{Z} \leftarrow \operatorname{NormMatch}(\mat{\widetilde C},\mat{C})$
\Comment{preserve conditioning magnitude}

\State \Return $\mat{Z}$
\end{algorithmic}
\end{algorithm}

\section{Experiments}
We first evaluate AdaIE on VLA models in both simulation and real-world manipulation settings. In simulation, we adopt RoboCasa-GR1~\citep{robocasa2024} and RoboTwin 2.0~\citep{chen2025robotwin}, following the settings of StarVLA~\citep{community2026starvla}. For real-world evaluation, we train and deploy the policy on IRON-R01-1.11 humanoid robots across diverse physical manipulation tasks. To examine whether AdaIE is limited to VLAs, we further extend it to world models, evaluating LeWM~\citep{maes2026leworldmodel} and DINO-WM~\citep{zhou2024dino} on multiple control tasks.

\subsection{Robotic Manipulation Benchmarks}
\subsubsection{Simulation Benchmarks}

Following StarVLA~\citep{community2026starvla}, we use identical training settings across variants, varying only $\alpha$. Full configurations are provided in Appendix~\ref{app:simulation_training_hyperparams}.

\paragraph{RoboCasa-GR1.}
RoboCasa-GR1~\citep{robocasa2024} is a tabletop manipulation benchmark built on the RoboCasa simulation framework and is commonly used to evaluate humanoid manipulation policies. The benchmark contains 24 tasks, with approximately 1,000 demonstrations per task, resulting in about 24K trajectories in total. We adopt a VLM--DiT architecture following GR00T-N1~\citep{bjorck2025gr00t}, with a pretrained Qwen3-VL backbone and a DiT action head from scratch. We train all policies with a batch size of 256 for 100K optimization steps. For evaluation, we run 60 rollouts per task.

As shown in Table~\ref{tab:robocasa_gr1}. AdaIE improves the baseline at both model scales. With the 2B backbone, AdaIE improves the success rate from $42.36\%$ to $47.64\%$, and achieves performance comparable to the results of 4B backbone. The improvement persists when scaling the backbone to 4B.

\begin{table}[t]
\centering
\caption{
Average success rate on RoboCasa-GR1 (24 tasks).
Results above the separator are reported by StarVLA~\citep{community2026starvla},
while results below are our runs, evaluated with 60 rollouts per task.
Detailed per-task results of our runs are provided in
Appendix~\ref{app:per_task_results}.
}
\begin{tabular}{lccc}
\toprule
\textbf{Method} & \textbf{Backbone} & \textbf{Steps} & \textbf{Avg. (\%)} \\
\midrule
$\pi_{0.5}$           & --           & --   & 37.00 \\
GR00T-N1.6            & --           & --   & 47.60 \\
StarVLA-GR00T         & Qwen3-VL-4B  & 100K & 47.80 \\
StarVLA-$\pi$         & Qwen3-VL-4B  & 100K & 43.90 \\
StarVLA-OFT           & Qwen3-VL-4B  & 100K & 48.80 \\
\midrule
GR00T                  & Qwen3-VL-2B & 100K & 42.36 \\
\rowcolor{blue!8}
GR00T + AdaIE           & Qwen3-VL-2B & 100K &
\textbf{47.64 \textcolor{green!50!black}{(+5.28)}} \\
GR00T                  & Qwen3-VL-4B & 100K & 47.99 \\
\rowcolor{blue!8}
GR00T + AdaIE           & Qwen3-VL-4B & 100K &
\textbf{50.07 \textcolor{green!50!black}{(+2.08)}} \\
\bottomrule
\end{tabular}
\label{tab:robocasa_gr1}
\end{table}

\paragraph{RoboTwin 2.0.}
RoboTwin 2.0~\citep{chen2025robotwin} is a large-scale benchmark for
bimanual robotic manipulation, covering 50 tasks under both clean and
randomized settings. Each task provides 50 clean and 500 randomized
demonstrations, resulting in approximately 27.5K trajectories in total.

Following the setup in StarVLA, we train the policy for 150K optimization steps with a global batch size of 192. For evaluation, we follow StarVLA and run 100 episodes per task, resulting in 10,000 evaluation episodes in total. The results are shown in Table~\ref{tab:robotwin2}.

\begin{table}[t]
\centering
\caption{
Success rates on RoboTwin 2.0 (50 tasks). Both methods are trained for 150K optimization steps using identical training configurations. See per task comparison in the Appendix~\ref{app:per_task_results}.}
\begin{tabular}{lccc}
\toprule
\textbf{Method}
& \textbf{Backbone}
& \textbf{Steps}
& \textbf{Success Rate (\%)} \\
\midrule
GR00T
& Cosmos-Reason2-2B
& 150K
& 66.92 \\
\rowcolor{blue!8}
GR00T + AdaIE
& Cosmos-Reason2-2B
& 150K
& \textbf{68.60\textcolor{green!50!black}{(+1.68)}} \\
\bottomrule
\end{tabular}
\label{tab:robotwin2}
\end{table}

\paragraph{Training overhead.}
AdaIE introduces only modest additional training cost. On the 2B model using
$4\times$H200 GPUs for 150K training steps, the baseline requires 65.45 hours,
while AdaIE requires 67.88 hours, corresponding to a 3.7\% increase in wall-clock training time.

\subsubsection{Real-world Robots Manipulation}
\label{sec:real_world}

To evaluate AdaIE under real physical interaction, we deploy the policy on an IRON-R01-1.11 dual-arm humanoid robot.
The policy receives a single head-mounted fisheye observation, a language instruction, and the current proprioceptive state, and predicts action chunks for manipulation.

\begin{figure}[h]
    \centering
    \includegraphics[width=\linewidth]{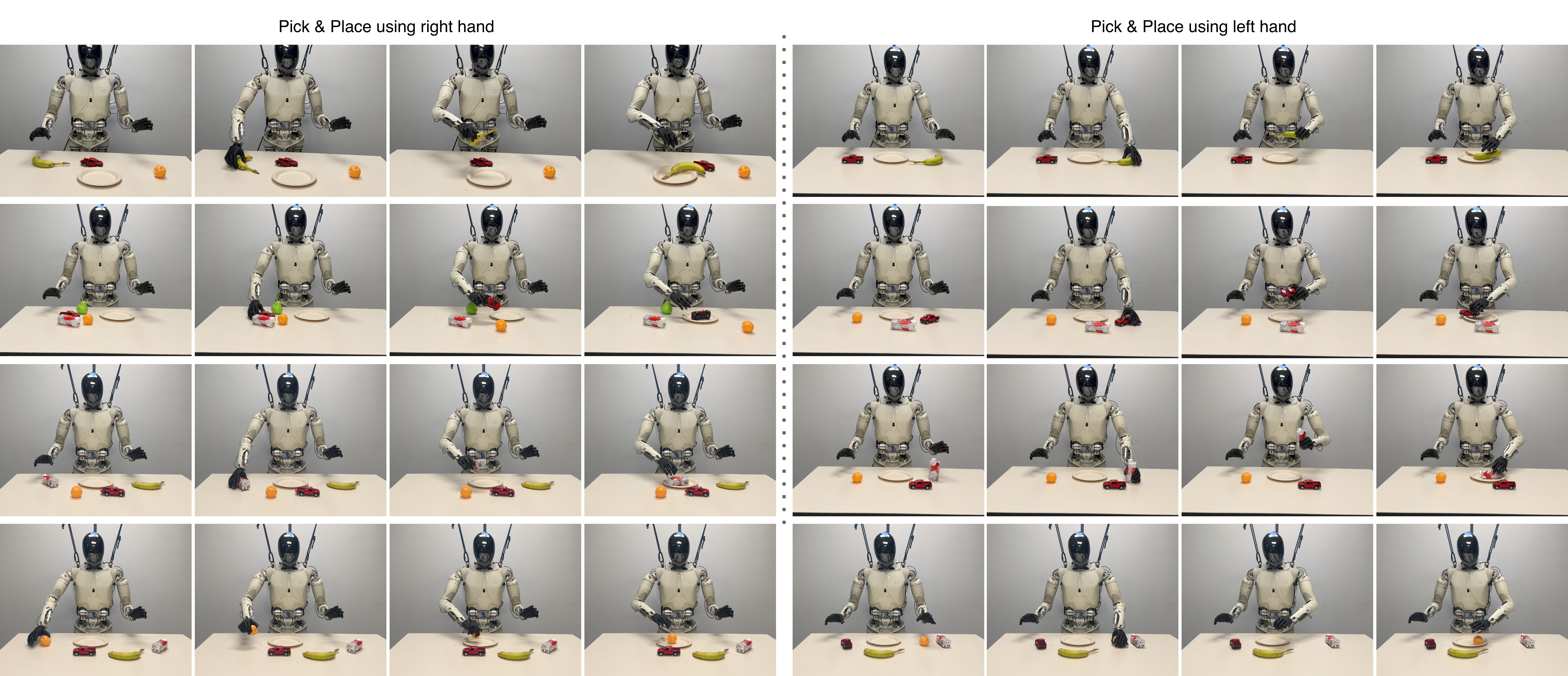}
    \caption{
    Real-world manipulation evaluation. For each task, we conduct 10 trials with the left hand and 10 trials with the right hand. See Table~\ref{tab:real_robot_detailed} for the detailed results.
    }
    \label{fig:placeholder}
\end{figure}

\paragraph{Data and Training.} We train on a curated real-robot dataset collected from five humanoid robots, comprising 5,875 teleoperated episodes across 21 manipulation task categories. The original 50-Hz trajectories are sampled at 10 Hz. Each training sample consists of an RGB observation, a language instruction, the current robot state, and a 30-step action chunk.

We adopt the same GR00T architecture as in simulation, using a pretrained Qwen3-VL-2B-Instruct backbone and a DiT action head trained from scratch. The baseline and AdaIE variants use exactly the same training data, optimization schedule, and model architecture, differing only by the insertion of AdaIE. Full dataset composition and training details are provided in Appendix~\ref{app:real_robot_details}.

\paragraph{Evaluation.}
We evaluate the baseline and AdaIE on 5 real-world manipulation tasks. Each method is evaluated for
20 trials (10 trials for left hand, and 10 for right hand) per task under identical initial conditions and task instructions. A trial is considered successful if the robot completes the specified task objective according to a predefined task-specific criterion, without human intervention. 
We additionally report {Interaction Success}, which measures whether the robot successfully reaches and interacts with the target object or mechanism, even if the full task objective is not completed. For example, in pick-and-place tasks, a trial is counted as an interaction success if the robot successfully grasps the target object but fails to place it in the designated tray. The results are shown in Table~\ref{tab:real_robot_detailed}, where AdaIE improves the overall task success rate compares to the baseline model.

\begin{table}[t]
\centering
\caption{
Real-robot evaluation results on in-distribution (ID) and
out-of-distribution (OOD) tasks.}
\small
\begin{tabular}{llcccc}
\toprule
& &
\multicolumn{2}{c}{\textbf{Baseline}}
& \multicolumn{2}{c}{\textbf{AdaIE}} \\
\cmidrule(lr){3-4}
\cmidrule(lr){5-6}
\textbf{Task}
& \textbf{Hand}
& \textbf{Task Succ.}
& \textbf{Interaction Succ.}
& \textbf{Task Succ.}
& \textbf{Interaction Succ.} \\
\midrule

\multicolumn{6}{l}{\textbf{In-Distribution (ID)}} \\
\addlinespace[2pt]

Banana
& Left  & \cellcolor{gray!10} 5/10 & \cellcolor{gray!10} 10/10 & \cellcolor{blue!8} \textbf{6/10} & \cellcolor{blue!8} 9/10 \\
& Right & \cellcolor{gray!10} 4/10 & \cellcolor{gray!10} 10/10 & \cellcolor{blue!8} \textbf{7/10} & \cellcolor{blue!8} 10/10 \\

Press Bell
& Left  & \cellcolor{gray!10} \textbf{6/10} & \cellcolor{gray!10} 9/10 & \cellcolor{blue!8} 5/10 & \cellcolor{blue!8} 10/10 \\
& Right & \cellcolor{gray!10} 2/10 & \cellcolor{gray!10} 9/10 & \cellcolor{blue!8} \textbf{4/10} & \cellcolor{blue!8} 8/10 \\

\addlinespace[2pt]
ID Overall
& -- & \cellcolor{gray!10} 42.5\% & \cellcolor{gray!10} \textbf{95.0\%}
& \cellcolor{blue!8} \textbf{55.0\%\textcolor{green!50!black}{(+12.5\%)}} & \cellcolor{blue!8} 92.5\% \textcolor{red!50!black}{(-2.5\%)} \\

\midrule

\multicolumn{6}{l}{\textbf{Out-of-Distribution (OOD)}} \\
\addlinespace[2pt]

Toy Car
& Left  & \cellcolor{gray!10} 4/10 & \cellcolor{gray!10} 10/10 & \cellcolor{blue!8} \textbf{5/10} & \cellcolor{blue!8} 10/10 \\
& Right & \cellcolor{gray!10} 2/10 & \cellcolor{gray!10} 9/10  & \cellcolor{blue!8} \textbf{5/10} & \cellcolor{blue!8} 10/10 \\

Cheese Ball
& Left  & \cellcolor{gray!10} 2/10 & \cellcolor{gray!10} 7/10 & \cellcolor{blue!8} \textbf{3/10} & \cellcolor{blue!8} 7/10 \\
& Right & \cellcolor{gray!10} 2/10 & \cellcolor{gray!10} 5/10 & \cellcolor{blue!8} \textbf{5/10} & \cellcolor{blue!8} 8/10 \\

Milk Bottle
& Left  & \cellcolor{gray!10} 2/10 & \cellcolor{gray!10} 7/10 & \cellcolor{blue!8} \textbf{5/10} & \cellcolor{blue!8} 8/10 \\
& Right & \cellcolor{gray!10} 1/10 & \cellcolor{gray!10} 8/10 & \cellcolor{blue!8} \textbf{4/10} & \cellcolor{blue!8} 10/10 \\

\addlinespace[2pt]
OOD Overall
& -- & \cellcolor{gray!10} 21.7\% & \cellcolor{gray!10} 76.7\%
& \cellcolor{blue!8} \textbf{45.0\%\textcolor{green!50!black}{(+23.3\%)}} & \cellcolor{blue!8} \textbf{88.3\%\textcolor{green!50!black}{(+11.7\%)}} \\

\midrule

\textbf{Overall}
& -- & \cellcolor{gray!10} 30.0\% & \cellcolor{gray!10} 84.0\%
& \cellcolor{blue!8} \textbf{49.0\%\textcolor{green!50!black}{(+19.0\%)}} & \cellcolor{blue!8} \textbf{90.0\%\textcolor{green!50!black}{(+6.0\%)}} \\

\bottomrule
\end{tabular}

\label{tab:real_robot_detailed}
\end{table}

\subsection{Extension to World Models}
\label{sec:adaie_wms}
AdaIE is not limited to VLA architectures and can be applied to more general cross-model interfaces. We further apply AdaIE to a JEPA-based~\citep{lecun2022path} world-model framework, using LeWM~\citep{maes2026leworldmodel} and DINO-WM\citep{zhou2024dino} as representative instantiations.

\paragraph{LeWorldModel.}
Following the experimental setup of LeWM~\citep{maes2026leworldmodel}, we train both the original LeWM baselines and their AdaIE variants on four tasks using identical configurations. LeWM is an end-to-end JEPA consisting of a ViT encoder that maps image observations into compact latent representations and a predictor that predicts future latent states. The encoder and predictor are jointly optimized with a latent prediction objective. We apply AdaIE to the encoder representation before it is consumed by the predictor. 

The LeWM evaluation contains four tasks: Two Room, Push-T, Reacher, and OGBench-Cube, covering 2D navigation, 2D manipulation and control, and 3D robotic manipulation. We run 5 random seed training for all variants for LeWM, and report mean$\pm$ standard deviation over five training seeds in Table~\ref{tab:lewm_results}. See Appendix~\ref{app:lewm_details} for additional experimental details.

\begin{figure}[t]
\centering

\begin{minipage}[c]{0.5\linewidth}
\centering
\small

\resizebox{\linewidth}{!}{
\begin{tabular}{llc}
\toprule
\textbf{Task} & \textbf{Method}
& \textbf{Success Rate (\%)} \\
\midrule

\multirow{3}{*}{Two Room}
& \cellcolor{gray!10} LeWM
& \cellcolor{gray!10} 91.6 $\pm$ 2.0 \\
& \cellcolor{gray!10} LeWM + Var.
& \cellcolor{gray!10}  91.2 $\pm$ 1.6 \\
& \cellcolor{blue!8} LeWM + AdaIE
& \cellcolor{blue!8} \textbf{93.2 $\pm$ 1.0} \\
\midrule

\multirow{3}{*}{Push-T}
& \cellcolor{gray!10} LeWM
& \cellcolor{gray!10} 92.8 $\pm$ 3.7 \\
& \cellcolor{gray!10} LeWM + Var.
& \cellcolor{gray!10}  93.6 $\pm$ 1.5 \\
& \cellcolor{blue!8} LeWM + AdaIE
& \cellcolor{blue!8} \textbf{95.6 $\pm$ 1.5} \\
\midrule

\multirow{3}{*}{Reacher}
& \cellcolor{gray!10} LeWM
& \cellcolor{gray!10} 76.4 $\pm$ 3.7 \\
& \cellcolor{gray!10} LeWM + Var.
& \cellcolor{gray!10}  76.8 $\pm$ 4.1 \\
& \cellcolor{blue!8} LeWM + AdaIE
& \cellcolor{blue!8} \textbf{ 78.4 $\pm$ 5.9} \\
\midrule

\multirow{3}{*}{OGBench-Cube}
& \cellcolor{gray!10} LeWM
& \cellcolor{gray!10}  76.4 $\pm$ 2.3\\
& \cellcolor{gray!10} LeWM + Var.
& \cellcolor{gray!10}   \textbf{77.2 $\pm$ 1.0} \\
& \cellcolor{blue!8} LeWM + AdaIE
& \cellcolor{blue!8} { 76.8 $\pm$ 3.2} \\
\bottomrule
\end{tabular}
}

\captionof{table}{
Planning performance of LeWM and LeWM + AdaIE.
The final planning success rate is reported. For each model, we follow the eval policy of LeWM~\citep{maes2026leworldmodel}, with 50 episodes per task. We use $\alpha=0.25$ for AdaIE.}
\label{tab:lewm_results}
\end{minipage}
\hfill
\begin{minipage}[c]{0.45\linewidth}
\centering

\includegraphics[
    width=\linewidth,
    trim=0 5 0 5,
    clip
]{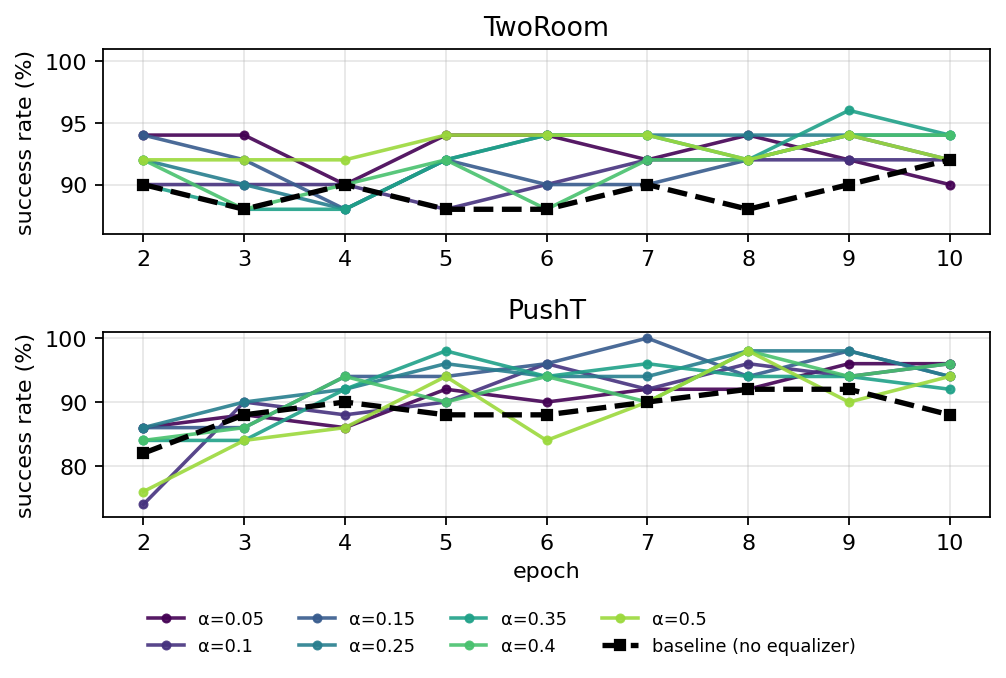}

\captionof{figure}{
Ablation of $\alpha\in\{0.05,\cdots,0.5\}$ on LeWM. 
We report evaluation success rates across training epochs. Full quantitative results are provided in Table~\ref{tab:lewm_alpha_full}.
}
\label{fig:ep_eval_s42}
\end{minipage}

\end{figure}

\paragraph{Ablation on $\alpha$}. We further study the effect of $\alpha$. As shown in Figure~\ref{fig:ep_eval_s42}, most $\alpha$ values improve planning performance over the baseline, although some settings are slightly worse than the baseline. The best $\alpha$ also differs across tasks, and how to derive the reweighting strength adaptively without task-specific tuning remains an open question. We provide the full per-epoch results in Table~\ref{tab:lewm_alpha_full}.

\paragraph{Sensitivity vs. variance.}
We further test whether downstream sensitivity can be replaced by the channel-wise variance of the pretrained representation. 
To test this, we replace the sensitivity estimate $\hat g_j$ with the channel variance $\hat\sigma_j^2$, while keeping the rest of the reweighting pipeline identical. 
As shown in Table~\ref{tab:lewm_results}, variance-based reweighting also yields gains on several tasks, while AdaIE performs the best in three of four over them.

\paragraph{DINO-WM.}
We further evaluate AdaIE on DINO-WM~\citep{zhou2024dino}. Following the DINO-WM setting adopted by LeWM~\citep{maes2026leworldmodel}, we train all models for 10 epochs. For evaluation, we follow the same model-predictive planning protocol as LeWM and report mean success rate over 10 random seeds, with 50 episodes per seed. The results are reported in Table~\ref{tab:dinowm_results}.
\begin{table}[t]

\caption{
Performance of DINO-WM and DINO-WM + AdaIE.
}
\centering
\footnotesize
\begin{tabular}{lcc@{\hspace{1.2em}}lcc}
\toprule
\multicolumn{3}{c}{\textbf{Push-T}}
&
\multicolumn{3}{c}{\textbf{Reacher}} \\
\cmidrule(lr){1-3}
\cmidrule(lr){4-6}
\textbf{Method}
& \textbf{Success (\%)}
& \textbf{Std.}
& \textbf{Method}
& \textbf{Success (\%)}
& \textbf{Std.} \\
\midrule

\cellcolor{gray!10} DINO-WM
& \cellcolor{gray!10} 71.4
& \cellcolor{gray!10} 7.0
&
\cellcolor{gray!10} DINO-WM
& \cellcolor{gray!10} 78.6
& \cellcolor{gray!10} 3.9 \\

\cellcolor{blue!8} DINO-WM + AdaIE
& \cellcolor{blue!8}
  \textbf{73.2 \textcolor{green!50!black}{(+1.8)}}
& \cellcolor{blue!8} 6.5
&
\cellcolor{blue!8} DINO-WM + AdaIE
& \cellcolor{blue!8}
  \textbf{82.8 \textcolor{green!50!black}{(+4.2)}}
& \cellcolor{blue!8} 6.3 \\

\bottomrule
\end{tabular}
\label{tab:dinowm_results}
\end{table}

\section{Related Work}

\subsection{Vision-Language-Action Models}

Vision-language-action (VLA) models integrate visual observations, language instructions, and robot actions into a unified policy. Early approaches such as RT-1 and RT-2~\citep{brohan2022rt,brohan2023rt} formulate action generation autoregressively, while more recent systems increasingly adopt continuous-action diffusion or flow-matching policies, including $\pi_0$~\citep{black2024pi_0} and GR00T N1~\citep{bjorck2025gr00t}. Open-source models such as Octo~\citep{mees2024octo} and OpenVLA~\citep{kim2024openvla} further demonstrate the effectiveness of transferring pretrained vision-language representations to robotic control. Many follow-up works improve these frameworks through action-model design~\citep{li2024cogact}, efficient fine-tuning and action generation~\citep{kim2025fine,pertsch2025fast}, and large-scale cross-task and cross-embodiment training~\citep{intelligence2025pi_}.

\subsection{World Models}

World models learn representations of environment dynamics to predict future states and support downstream decision-making and planning. Early approaches learn latent dynamics for model-based control~\citep{ha2018world,hafner2019dream}, recent methods explore predictive representation learning, such as JEPA~\citep{lecun2022path}, and pretrained visual representations for dynamics prediction. DINO-WM~\citep{zhou2024dino} and LeWM~\citep{maes2026leworldmodel}, for example, predict future states in representation space and use the learned dynamics for downstream planning. Many other following works explore the optimization of those frameworks, including training~\citep{agarwal2026cosmos,kang2026x}, planning~\citep{psenka2026parallel} and others~\citep{zhang2026hierarchical,klindt2026does,luo2026toward,rao2026skyjepa}.

\section{Conclusion}

We uncover \emph{interface co-adaptation}: downstream models progressively adapt their sensitivity to the anisotropic geometry of the backbone representations. Our observations suggest that counteracting sensitivity concentration can benefit downstream learning. We introduce AdaIE to regulate it through sensitivity-based reweighting. Across VLAs, diffusion policies, and world models, AdaIE consistently improves downstream performance. More broadly, our results establish the representation interface itself as a important optimization problem in modular learning systems.

\clearpage

\bibliography{iclr2027_conference}
\bibliographystyle{iclr2027_conference}
\clearpage
\appendix

\section{Appendix}

\subsection{Training Configuration on Robocasa-GR1/Robotwin 2.0}
\label{app:simulation_training_hyperparams}
\begin{table}[h]
\centering
\small
\begin{tabular}{lcc}
\toprule
\textbf{Configuration} & \textbf{GR1} & \textbf{RoboTwin 2.0} \\
\midrule
VLM backbone & Qwen3-VL-2B-Instruct & Cosmos-Reason2-2B \\
Action head & 28-layer DiT, 1024-d & 28-layer DiT, 1024-d \\
Training steps & 100K & 150K \\
Effective batch size & 256 & 192 \\
Optimizer & AdamW & AdamW \\
VLM learning rate & $1\times10^{-5}$ & $1\times10^{-5}$ \\
DiT learning rate & $1\times10^{-4}$ & $1\times10^{-4}$ \\
LR schedule & Warmup + cosine & Warmup + cosine \\
Precision & BF16 & BF16 \\
Action horizon & 40 & 40 \\
\midrule
AdaIE (treated variant) & \checkmark & \checkmark \\
$\alpha$ & 0.5 & 0.5 \\
Sensitivity EMA & 0.99 & 0.99 \\
Probe interval & 8 steps & 8 steps \\
Norm matching & RMS-preserving & RMS-preserving \\
\bottomrule
\end{tabular}
\caption{
Training configurations for the GR1 and RoboTwin~2.0 experiments. Within
each benchmark, the baseline and AdaIE models use identical architectures,
data, and optimization settings, differing only by the insertion of AdaIE
at the VLM--DiT interface (bottom block).
}
\label{tab:sim_training}
\end{table}

\paragraph{More Details of Norm matching.}
To prevent sensitivity reweighting from changing the overall feature magnitude,
we apply a group-wise norm-matching step after reweighting. For each group, we
maintain exponential moving averages of the input and output mean-square
magnitudes,
\begin{equation}
    m_{\mathrm{in}}
    \leftarrow
    \operatorname{EMA}_{\beta}\!\left[
    \operatorname{mean}(C^2)
    \right],
    \qquad
    m_{\mathrm{out}}
    \leftarrow
    \operatorname{EMA}_{\beta}\!\left[
    \operatorname{mean}(\widetilde C^2)
    \right],
\end{equation}
where the mean is computed over all valid tokens and feature dimensions, and
$\widetilde C=C\odot s$ denotes the reweighted representation before
normalization. We then rescale the output by
\begin{equation}
    r =
    \sqrt{
        \frac{m_{\mathrm{in}}}
        {\max(m_{\mathrm{out}},10^{-12})}
    },
    \qquad
    Z = r\,\widetilde C.
\end{equation}
All statistics and the rescaling factor are computed without gradient tracking.
This normalization preserves the relative channel-wise reweighting introduced
by AdaIE while matching the overall RMS magnitude of the reweighted
representation to that of the original input.

\subsection{Real-Robot Experimental Details}
\label{app:real_robot_details}
\paragraph{Robot Platform.}
Experiments are conducted on the IRON-R01-1.11 dual-arm humanoid platform. The robot is operated as a fixed-base manipulator, with two 7-DoF arms and dexterous hands. Visual observations are captured using the head-mounted fisheye camera. The policy uses a single camera stream together with proprioceptive observations of the upper body.

\begin{figure}[H]
    \centering
    \includegraphics[width=\linewidth]{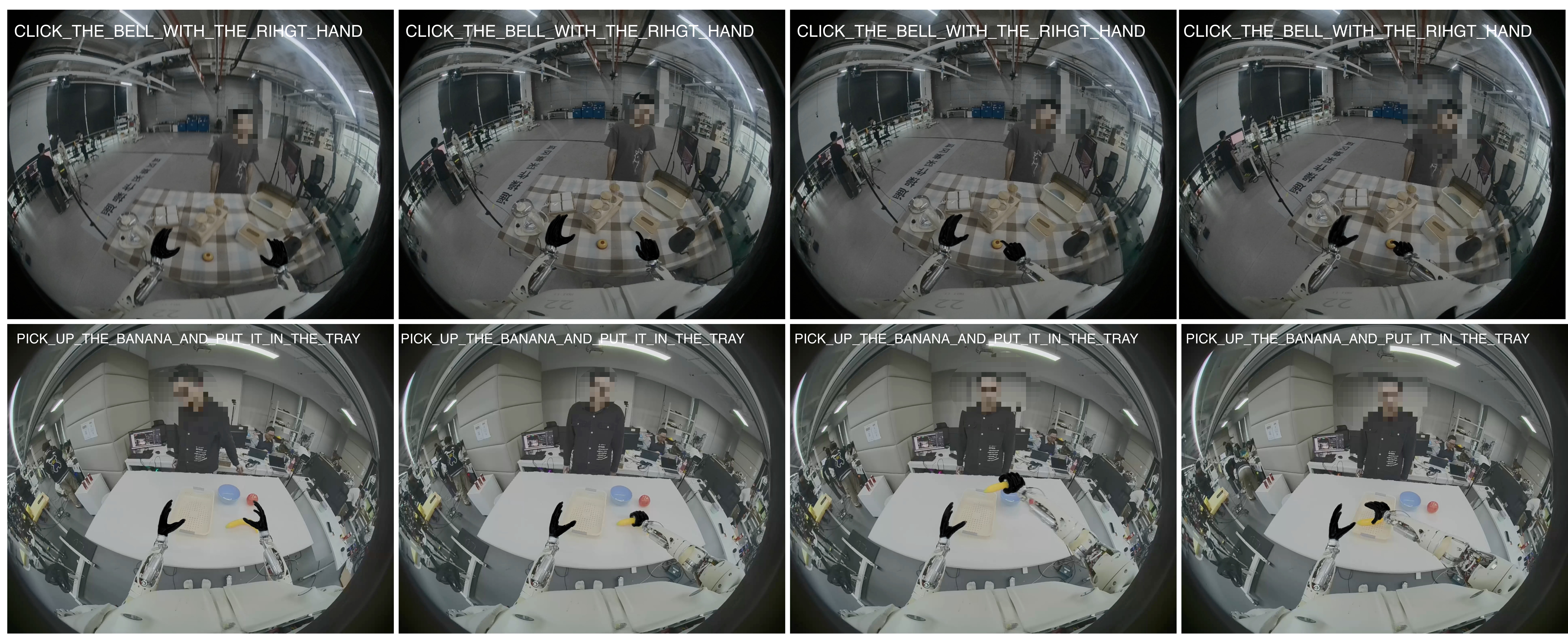}
    \caption{Representative examples from the real-robot training dataset.}
    \label{fig:rb_data}
\end{figure}
\paragraph{Dataset.}
The complete real-robot collection pool contains 37,362 episodes and approximately 16.8M frames collected from five R01 robots. From this pool, we construct a quality-filtered and approximately
task-balanced training subset containing 4,304 episodes (13.7 hours) across
21 task categories. We additionally include 1,571 laboratory pick-and-place
episodes (2.7 hours), resulting in 5,875 episodes and 16.4 hours of training data in total. Figure~\ref{fig:rb_data} shows representative training examples, with the data anonymized for privacy.

The original trajectories are recorded at 50 Hz and sampled at 10 Hz for
training. Each sample contains a head-mounted fisheye observation, a language
instruction, the current proprioceptive state, and a 30-step action chunk,
corresponding to a 3-s prediction horizon. Images are cropped and resized to
$320\times480$. We use an episode-level held-out validation split and compute
normalization statistics from the training corpus.

\paragraph{Model and Optimization.}
Our real-robot policy follows a GR00T architecture. A pretrained
Qwen3-VL-Instruct backbone encodes the visual observation and language
instruction, while a 28-layer DiT action head is initialized from scratch and
trained using flow matching. AdaIE is inserted between the VLM representation
and the action DiT. It introduces no additional trainable parameters.

All baseline and AdaIE variants use identical optimization configurations.
We use AdamW with BF16 training and distributed data parallelism on H200 GPUs.
The pretrained VLM backbone and the from-scratch DiT action head use learning
rates of $1\times10^{-5}$ and $1\times10^{-4}$, respectively. For AdaIE, we
set $\alpha=0.25$. All policies are trained for approximately 32K optimizer
steps with an effective batch size of 512. We compare models at matched
training exposure (checkpoint at step 32K) across all
variants. The complete training configuration is summarized in
Table~\ref{tab:real_robot_training}.

\begin{table}[h]
\centering
\small
\begin{tabular}{lcc}
\toprule
\textbf{Configuration} & \textbf{Baseline} & \textbf{AdaIE} \\
\midrule
VLM backbone & Qwen3-VL-2B-Instruct & Qwen3-VL-2B-Instruct \\
Action head & 28-layer DiT, 1024-d & 28-layer DiT, 1024-d \\
Training steps & 32K & 32K \\
Effective batch size & 512 & 512 \\
Optimizer &  AdamW &  AdamW \\
VLM learning rate & $1\times10^{-5}$ & $1\times10^{-5}$ \\
DiT learning rate & $1\times10^{-4}$ & $1\times10^{-4}$ \\
LR schedule & Warmup + cosine restarts & Warmup + cosine restarts \\
Precision & BF16 & BF16 \\
Action horizon & 30 & 30 \\
Flow-matching steps & 4 & 4 \\
\midrule
AdaIE & -- & \checkmark \\
$\alpha$ & -- & 0.25 \\
Sensitivity EMA & -- & 0.99 \\
Probe interval & -- & 8 steps \\
Norm matching & -- & RMS-preserving \\
\bottomrule
\end{tabular}
\caption{
Training configurations for the real-robot baseline and AdaIE models.
Both variants use identical model architectures, data, and optimization
settings, differing only by the insertion of AdaIE at the VLM--DiT interface.
}
\label{tab:real_robot_training}
\end{table}

\paragraph{Closed-Loop Inference.}
At deployment, the policy predicts 30 action waypoints per inference call.
The first two waypoints are discarded by the deployment controller, and five
waypoints are executed before replanning. Each waypoint has a duration of
200\,ms. The robot therefore repeatedly executes a short action segment,
acquires a new observation, and replans in a receding-horizon fashion.

The policy outputs torso-frame end-effector targets together with joint and
hand commands. These commands are passed to the robot's low-level whole-body
control stack, which performs inverse-kinematics and safety-constrained
execution. The perception, action-conversion, low-level control, and inference
settings are held fixed across all compared variants.

\clearpage

\subsection{Training Configuration on World Models}
\label{app:lewm_details}

\begin{table}[H]
\centering
\caption{
Training configurations for the LeWM and DINO-WM experiments. We use pre-trained backbone for two world model, with Tiny-ViT for LeWM and DINOv2 for DINO-WM.
}
\small
\begin{tabular}{lcc}
\toprule
\textbf{Configuration} & \textbf{LeWM} & \textbf{DINO-WM} \\
\midrule
Predictor & 6-block AdaLN-Zero DiT, 192-d & 6-layer ViT, 16 heads, MLP 2048 ($\sim$19M) \\
Training loss & next-emb.\ MSE + $0.09\,$SIGReg & latent next-state MSE (DINOv2 space) \\
Proprioception & unused & excluded (pixels + actions only) \\
Epochs & 10 & 10 \\
Batch size & 128 & 32 \\
Optimizer & AdamW & AdamW \\
Encoder learning rate & $1\times10^{-5}$ (joint)$^{\dagger}$ & --- (frozen) \\
Predictor learning rate & $5\times10^{-5}$ & $5\times10^{-5}$ \\
Weight decay & $1\times10^{-3}$ & 0 \\
LR schedule & 1\% warmup + cosine & constant \\
Gradient clipping & 1.0 & 1.0 \\
Precision & BF16 & BF16 \\
Image size / frameskip & $224$ / 5 & $224$ / 5 \\
History length & 3 (TwoRoom: 1) & 3 (TwoRoom: 1) \\
\midrule
AdaIE (treated variants) & \checkmark & \checkmark \\
Sensitivity EMA & 0.99 & 0.99 \\
Probe interval & 1 step & 1 step \\
Norm matching & RMS-preserving & RMS-preserving \\
\bottomrule
\end{tabular}
\label{tab:wm_training}
\end{table}

\clearpage

\subsection{Training Configuration of Section~\ref{sec:probe}}
\label{app:prob_detals}

\subsubsection{Depth Estimation on NYUv2}

\begin{table}[H]
\centering
\caption{
Training configuration for the DINOv2-S + DPT experiments on NYUv2, following the official Depth Anything V2 metric-depth recipe (SiLog loss, AdamW, polynomial decay, $518\times518$ crops) with the \texttt{train.py} default of 40 epochs and a global batch of 16. The baseline and AdaIE models are identical except for the insertion of AdaIE at the encoder--decoder interface (bottom block). The standard deviation reflects variation across 3 random seed training in Table~\ref{tab:aligned_vs_mismatch}.
}
\footnotesize
\begin{tabular}{lc}
\toprule
\textbf{Configuration} & \textbf{DINOv2-S + DPT / NYUv2} \\
\midrule
Task / dataset & NYU Depth V2, BTS/Eigen split (24{,}231 train pairs, 654 test images) \\
Input & $480\times640$ RGB, ImageNet mean/std \\
Resize / crop (train / eval) & short side to 518 (multiple of 14) + random $518\times518$ crop / resize only \\
Depth target & 16-bit PNG / 1000 (m), valid range $[0.001, 10]$ m \\
Backbone & DINOv2 ViT-S/14, 12 blocks, width 384 \\
Backbone init & DINOv2 self-supervised weights (\texttt{dinov2\_vits14\_pretrain}) \\
Decoder & DPT head (features 64, channels 48/96/192/384), random init \\
Decoder input & patch tokens from blocks $\{2, 5, 8, 11\}$ after final LayerNorm (no CLS readout) \\
Output & sigmoid $\times$ max depth (10 m), bilinear upsampling to $480\times640$ at eval \\
Augmentation & horizontal flip ($p=0.5$) \\
Loss & scale-invariant log (SiLog), $\lambda=0.5$ \\
Training epochs & 40 \\
Batch size & 16 (4 GPUs $\times$ 4) \\
Optimizer & AdamW \\
Learning rate (decoder) & $5\times10^{-5}$ \\
Backbone learning rate$^{\dagger}$ & $5\times10^{-6}$ \\
Betas & $(0.9, 0.999)$ \\
Weight decay & $0.01$ \\
LR schedule & polynomial decay $(1 - t/T)^{0.9}$, no warmup \\
Gradient clipping & none \\
Precision & float32 \\
EMA & none \\
Evaluation & 654 test images every epoch, Eigen crop, depth capped at 10 m; final epoch reported \\
Metrics & $\delta_1, \delta_2, \delta_3$, AbsRel, RMSE, $\log_{10}$ \\
\midrule
AdaIE (treated variants) & \checkmark \\
Insertion point & encoder taps $\to$ DPT decoder (input side) \\
Groups & 4 $\times$ 384 (one per tap) \\
$\alpha$ &  0.5/-0.5 \\
Sensitivity EMA & 0.99 \\
Probe interval & 1 step \\
Norm matching & RMS-preserving \\
\bottomrule
\end{tabular}
\label{tab:dpt_nyu_training}
\end{table}

\clearpage

\subsubsection{ViT Classification on CIFAR-100}

\begin{table}[H]
\centering
\caption{
Training configuration for the DeiT-S experiments on CIFAR-100. The schedule follows the ViT fine-tuning budget (10k steps, batch 512) with DeiT AdamW defaults. The baseline and AdaIE models are identical except for the insertion of AdaIE at the backbone--classifier interface. The standard deviation reflects variation across the 5 random seed training in Table~\ref{tab:aligned_vs_mismatch}.
}
\footnotesize
\begin{tabular}{lc}
\toprule
\textbf{Configuration} & \textbf{DeiT-S / CIFAR-100} \\
\midrule
Task / dataset & CIFAR-100 (49k train, 1k minival held out from train, 10k test) \\
Input & $32\times32$ RGB, bicubic upsampling to $224\times224$, ImageNet mean/std \\
Backbone & DeiT-S (\texttt{deit\_small\_patch16\_224}, timm), 12 blocks, width 384, patch 16 \\
Backbone init & ImageNet-1k supervised DeiT weights \\
Classifier head & Linear $384\to100$, random init (trunc.\ normal, std 0.02; zero bias) \\
Head input & CLS token after final LayerNorm (pre-logits) \\
Augmentation (train / eval) & random crop (4-px pad) + horizontal flip at $32\times32$ / none \\
Mixup / CutMix / RandAugment & none \\
Label smoothing / dropout & 0 / 0 \\
Stochastic depth & 0.1 \\
Training steps & 10{,}000 ($\approx$105 epochs) \\
Batch size & 512 \\
Optimizer & AdamW \\
Learning rate (head) & $1\times10^{-3}$ \\
Backbone learning rate$^{\dagger}$ & $1\times10^{-4}$ \\
Betas & $(0.9, 0.999)$ \\
Weight decay & $0.05$ (none on biases, norms, pos.\ embed, CLS token) \\
LR schedule & cosine to $1\times10^{-6}$, 500 warmup steps \\
Gradient clipping & global norm 1.0 \\
Precision & bfloat16 autocast \\
EMA & none \\
\midrule
AdaIE (treated variants) & \checkmark \\
Insertion point & pre-logits $\to$ classifier head\\
$\alpha$ & 0.5/-0.5 \\
Sensitivity EMA & 0.99 \\
Probe interval & 1 step \\
Norm matching & RMS-preserving \\
\bottomrule
\end{tabular}
\label{tab:deit_cifar_training}
\end{table}

\clearpage

\subsubsection{Diffusion Policy on Push-T}

\begin{table}[H]
\centering
\caption{
Training configuration for the DiffusionPolicy-C experiments on Push-T, following the official Diffusion Policy recipe. The baseline and AdaIE use identical configurations except for the insertion of AdaIE at the observation--denoiser interface. Following the original evaluation protocol, we report the best checkpoint performance for each variant as the mean and standard deviation of test scores across three random seeds.
}
\footnotesize
\begin{tabular}{lc}
\toprule
\textbf{Configuration} & \textbf{DiffusionPolicy-C} \\
\midrule
Task / dataset & PushT image (\texttt{pusht\_cchi\_v7\_replay}, 90 training episodes, 2\% val) \\
Observation & $96\times96$ RGB + 2-D agent position, $T_o = 2$ \\
Vision encoder & Pre-trained ResNet-18, GroupNorm, spatial-softmax pooling \\
Random crop (train / eval) & $84\times84$ / fixed \\
Denoiser & 1-D convolutional UNet (512, 1024, 2048) \\
Conditioning & FiLM (scale + shift), diffusion-step embed 128 \\
Diffusion & DDPM, 100 train / 100 inference steps, squared-cosine $\beta$, $\epsilon$-prediction \\
Action horizon / executed & 16 / 8 \\
Training epochs & 3050 \\
Batch size & 64 \\
Optimizer & AdamW \\
Learning rate & $1\times10^{-4}$ \\
Betas & $(0.95, 0.999)$ \\
Weight decay & $1\times10^{-6}$ \\
Backbone learning rate$^{\dagger}$ & $1\times10^{-5}$ \\
LR schedule & cosine, 500 warmup steps \\
EMA & power 0.75, max decay 0.9999 \\
Evaluation & 50 episodes every 50 epochs (max 300 steps); \\
\midrule
AdaIE (treated variants) & \checkmark \\
Insertion point & encoder features $\to$ denoiser conditioning (input side) \\
$\alpha$ & 0.25/-0.25 \\
Sensitivity EMA & 0.99 \\
Probe interval & 1 step \\
Norm matching & RMS-preserving \\
\bottomrule
\end{tabular}
\label{tab:dpc_training}
\end{table}

\subsubsection{VLM--DiT on RoboCasa-GR1}
 All compared variants use the same optimization configuration. We train with AdamW using $\beta_1=0.9$, $\beta_2=0.999$, weight decay $10^{-5}$, and gradient clipping at $1.0$. The learning rate follows a cosine schedule with 5\% warmup. We use a learning rate of $10^{-4}$ for the action DiT and $10^{-5}$ for the VLM backbone. We trained 30K steps for all variants in this experiment.
 
 For sensitivity estimation, we use one Hutchinson-style probe every 8 optimization steps and an EMA decay of $0.99$.  For the controlled experiment in Section~\ref{sec:probe}, the three variants differ only in the exponent $\alpha$:
with $\alpha=+\frac{1}{2}$ for \emph{Equalize}, $\alpha=0$ for the
\emph{Baseline}, and $\alpha=-\frac{1}{2}$ for \emph{Exploit}.

\paragraph{Tasks.}
The diagnostic study uses 12 RoboCasa GR1-tabletop tasks, with 1,000 demonstrations per task. The set contains 6 articulated-receptacle pick-and-place tasks, where the policy places an object into a cabinet, drawer, or microwave and then closes the receptacle:
\texttt{PnPBottleToCabinetClose},\\
\texttt{PnPCanToDrawerClose},\\
\texttt{PnPCupToDrawerClose},\\
\texttt{PnPMilkToMicrowaveClose},\\
\texttt{PnPPotatoToMicrowaveClose}, and \\
\texttt{PnPWineToCabinetClose}.\\
The remaining 6 tasks are novel-object surface-to-receptacle transfers
from the SplitA subset:
Cuttingboard$\rightarrow$Basket,
Cuttingboard$\rightarrow$Pot,
Placemat$\rightarrow$Bowl,
Placemat$\rightarrow$Tieredshelf,
Plate$\rightarrow$Pan, and
Tray$\rightarrow$Cardboardbox. We use 60 rollouts per task in the evaluation. The standard deviation reflects variation across the 5 random seed training in Table~\ref{tab:aligned_vs_mismatch}.

\clearpage

\subsection{Detailed Evaluation Results}
\label{app:per_task_results}

\begin{table*}[t]
\centering
\caption{
Per-task success rates on RoboCasa-GR1 for our GR00T baselines and AdaIE. 60 rollouts per task.
}
\label{tab:robocasa_gr1_per_task}
\resizebox{\textwidth}{!}{
\begin{tabular}{lcc|cc}
\toprule
\textbf{Task}
& \textbf{GR00T-2B}
& \textbf{GR00T-2B + AdaIE}
& \textbf{GR00T-4B}
& \textbf{GR00T-4B + AdaIE} \\
\midrule

PnPBottleToCabinetClose
& 50.00 & \textbf{56.67} & \textbf{65.00} & 48.33 \\

PnPCanToDrawerClose
& 65.00 & \textbf{78.33} & \textbf{80.00} & 75.00 \\

PnPCupToDrawerClose
& \textbf{23.33} & \textbf{23.33} & \textbf{23.33} & 15.00 \\

PnPMilkToMicrowaveClose
& \textbf{53.33} & 51.67 & \textbf{60.00} & \textbf{60.00} \\

PnPPotatoToMicrowaveClose
& 26.67 & \textbf{33.33} & 41.67 & \textbf{43.33} \\

PnPWineToCabinetClose
& \textbf{48.33} & \textbf{48.33} & \textbf{51.67} & \textbf{51.67} \\

\midrule

PnPNovelFromCuttingboardToBasket
& \textbf{46.67} & 45.00 & 56.67 & \textbf{61.67} \\

PnPNovelFromCuttingboardToCardboardbox
& 38.33 & \textbf{51.67} & 33.33 & \textbf{41.67} \\

PnPNovelFromCuttingboardToPan
& 56.67 & \textbf{58.33} & \textbf{56.67} & 55.00 \\

PnPNovelFromCuttingboardToPot
& 45.00 & \textbf{53.33} & \textbf{56.67} & 53.33 \\

PnPNovelFromCuttingboardToTieredbasket
& 40.00 & \textbf{46.67} & 43.33 & \textbf{45.00} \\

\midrule

PnPNovelFromPlacematToBasket
& 40.00 & \textbf{45.00} & 46.67 & \textbf{65.00} \\

PnPNovelFromPlacematToBowl
& 46.67 & \textbf{58.33} & 41.67 & \textbf{55.00} \\

PnPNovelFromPlacematToPlate
& 45.00 & \textbf{56.67} & 50.00 & \textbf{63.33} \\

PnPNovelFromPlacematToTieredshelf
& \textbf{26.67} & 11.67 & \textbf{31.67} & 26.67 \\

\midrule

PnPNovelFromPlateToBowl
& \textbf{45.00} & \textbf{45.00} & 61.67 & \textbf{65.00} \\

PnPNovelFromPlateToCardboardbox
& 33.33 & \textbf{41.67} & 36.67 & \textbf{40.00} \\

PnPNovelFromPlateToPan
& 31.67 & \textbf{60.00} & \textbf{43.33} & \textbf{43.33} \\

PnPNovelFromPlateToPlate
& 43.33 & \textbf{63.33} & 55.00 & \textbf{61.67} \\

\midrule

PnPNovelFromTrayToCardboardbox
& 43.33 & \textbf{46.67} & 38.33 & \textbf{50.00} \\

PnPNovelFromTrayToPlate
& 53.33 & \textbf{58.33} & 48.33 & \textbf{63.33} \\

PnPNovelFromTrayToPot
& \textbf{55.00} & 51.67 & 53.33 & \textbf{58.33} \\

PnPNovelFromTrayToTieredbasket
& \textbf{45.00} & 40.00 & \textbf{48.33} & 43.33 \\

PnPNovelFromTrayToTieredshelf
& 15.00 & \textbf{18.33} & \textbf{28.33} & 16.67 \\

\midrule
\textbf{Average}
& 42.36
& \textbf{47.64}
& 47.99
& \textbf{50.07} \\
\bottomrule
\end{tabular}
}
\end{table*}

\clearpage

\begin{table*}[t]
\centering
\caption{
Per-task success rates on RoboTwin 2.0 for the GR00T-2B baseline and AdaIE, evaluated with 100 rollouts per task per setup.
Clean uses the standard environment, while Randomized introduces variations in backgrounds, lighting, distractors, tabletop height, and language instructions.
Avg denotes the mean over both setups.
AdaIE improves average performance primarily under Clean.
The limited gain under Randomized may reflect additional perceptual and generalization demands that are not directly addressed by interface sensitivity reweighting.
}

\label{tab:robotwin_per_task}
\resizebox{\textwidth}{!}{
\begin{tabular}{lcc|cc|cc}
\toprule
\textbf{Task}
& \multicolumn{2}{c|}{\textbf{Clean}}
& \multicolumn{2}{c|}{\textbf{Randomized}}
& \multicolumn{2}{c}{\textbf{Avg}} \\
& \textbf{GR00T-2B} & \textbf{+ AdaIE}
& \textbf{GR00T-2B} & \textbf{+ AdaIE}
& \textbf{GR00T-2B} & \textbf{+ AdaIE} \\
\midrule

adjust\_bottle
& 94.00 & \textbf{99.00} & 97.00 & \textbf{98.00} & 95.50 & \textbf{98.50} \\

beat\_block\_hammer
& 62.00 & \textbf{76.00} & \textbf{74.00} & \textbf{74.00} & 68.00 & \textbf{75.00} \\

blocks\_ranking\_rgb
& 65.00 & \textbf{71.00} & \textbf{68.00} & 64.00 & 66.50 & \textbf{67.50} \\

blocks\_ranking\_size
& \textbf{60.00} & 47.00 & \textbf{54.00} & 46.00 & \textbf{57.00} & 46.50 \\

click\_alarmclock
& \textbf{100.00} & 99.00 & \textbf{100.00} & \textbf{100.00} & \textbf{100.00} & 99.50 \\

click\_bell
& \textbf{100.00} & 99.00 & \textbf{100.00} & \textbf{100.00} & \textbf{100.00} & 99.50 \\

dump\_bin\_bigbin
& 90.00 & \textbf{92.00} & \textbf{94.00} & 92.00 & \textbf{92.00} & \textbf{92.00} \\

grab\_roller
& 97.00 & \textbf{98.00} & \textbf{98.00} & 95.00 & \textbf{97.50} & 96.50 \\

handover\_block
& \textbf{74.00} & 65.00 & 54.00 & \textbf{61.00} & \textbf{64.00} & 63.00 \\

handover\_mic
& 80.00 & \textbf{93.00} & 87.00 & \textbf{92.00} & 83.50 & \textbf{92.50} \\

hanging\_mug
& \textbf{19.00} & 4.00 & \textbf{17.00} & 10.00 & \textbf{18.00} & 7.00 \\

lift\_pot
& 45.00 & \textbf{69.00} & 45.00 & \textbf{67.00} & 45.00 & \textbf{68.00} \\

move\_can\_pot
& 46.00 & \textbf{48.00} & \textbf{40.00} & 37.00 & \textbf{43.00} & 42.50 \\

move\_pillbottle\_pad
& 63.00 & \textbf{71.00} & 58.00 & \textbf{65.00} & 60.50 & \textbf{68.00} \\

move\_playingcard\_away
& 77.00 & \textbf{94.00} & 82.00 & \textbf{93.00} & 79.50 & \textbf{93.50} \\

move\_stapler\_pad
& 23.00 & \textbf{30.00} & 27.00 & \textbf{39.00} & 25.00 & \textbf{34.50} \\

open\_laptop
& 81.00 & \textbf{84.00} & \textbf{95.00} & 92.00 & \textbf{88.00} & \textbf{88.00} \\

open\_microwave
& 36.00 & \textbf{37.00} & \textbf{46.00} & 43.00 & \textbf{41.00} & 40.00 \\

pick\_diverse\_bottles
& \textbf{77.00} & 73.00 & \textbf{67.00} & 49.00 & \textbf{72.00} & 61.00 \\

pick\_dual\_bottles
& \textbf{94.00} & 78.00 & \textbf{88.00} & 62.00 & \textbf{91.00} & 70.00 \\

place\_a2b\_left
& 45.00 & \textbf{56.00} & 55.00 & \textbf{66.00} & 50.00 & \textbf{61.00} \\

place\_a2b\_right
& \textbf{58.00} & 57.00 & 49.00 & \textbf{58.00} & 53.50 & \textbf{57.50} \\

place\_bread\_basket
& 77.00 & \textbf{83.00} & \textbf{83.00} & 79.00 & 80.00 & \textbf{81.00} \\

place\_bread\_skillet
& 83.00 & \textbf{87.00} & \textbf{81.00} & 78.00 & 82.00 & \textbf{82.50} \\

place\_burger\_fries
& 82.00 & \textbf{92.00} & \textbf{92.00} & 89.00 & 87.00 & \textbf{90.50} \\

\midrule

place\_can\_basket
& \textbf{38.00} & 36.00 & \textbf{42.00} & 40.00 & \textbf{40.00} & 38.00 \\

place\_cans\_plasticbox
& 27.00 & \textbf{78.00} & 33.00 & \textbf{60.00} & 30.00 & \textbf{69.00} \\

place\_container\_plate
& 93.00 & \textbf{94.00} & 85.00 & \textbf{89.00} & 89.00 & \textbf{91.50} \\

place\_dual\_shoes
& \textbf{52.00} & 51.00 & 57.00 & \textbf{70.00} & 54.50 & \textbf{60.50} \\

place\_empty\_cup
& 95.00 & \textbf{98.00} & \textbf{95.00} & 86.00 & \textbf{95.00} & 92.00 \\

place\_fan
& 42.00 & \textbf{49.00} & \textbf{53.00} & 45.00 & \textbf{47.50} & 47.00 \\

place\_mouse\_pad
& 29.00 & \textbf{42.00} & 31.00 & \textbf{46.00} & 30.00 & \textbf{44.00} \\

place\_object\_basket
& 73.00 & \textbf{84.00} & 79.00 & \textbf{81.00} & 76.00 & \textbf{82.50} \\

place\_object\_scale
& 64.00 & \textbf{71.00} & 61.00 & \textbf{66.00} & 62.50 & \textbf{68.50} \\

place\_object\_stand
& \textbf{81.00} & \textbf{81.00} & \textbf{82.00} & 81.00 & \textbf{81.50} & 81.00 \\

place\_phone\_stand
& 76.00 & \textbf{80.00} & 78.00 & \textbf{80.00} & 77.00 & \textbf{80.00} \\

place\_shoe
& \textbf{87.00} & 75.00 & 78.00 & \textbf{79.00} & \textbf{82.50} & 77.00 \\

press\_stapler
& 86.00 & \textbf{89.00} & \textbf{89.00} & 87.00 & 87.50 & \textbf{88.00} \\

put\_bottles\_dustbin
& 37.00 & \textbf{53.00} & \textbf{56.00} & 39.00 & \textbf{46.50} & 46.00 \\

put\_object\_cabinet
& \textbf{51.00} & 41.00 & \textbf{41.00} & 31.00 & \textbf{46.00} & 36.00 \\

rotate\_qrcode
& 76.00 & \textbf{78.00} & 71.00 & \textbf{75.00} & 73.50 & \textbf{76.50} \\

scan\_object
& \textbf{76.00} & 53.00 & \textbf{59.00} & 54.00 & \textbf{67.50} & 53.50 \\

shake\_bottle
& \textbf{97.00} & 96.00 & \textbf{100.00} & 96.00 & \textbf{98.50} & 96.00 \\

shake\_bottle\_horizontally
& 96.00 & \textbf{97.00} & \textbf{99.00} & 95.00 & \textbf{97.50} & 96.00 \\

stack\_blocks\_three
& 17.00 & \textbf{32.00} & 22.00 & \textbf{29.00} & 19.50 & \textbf{30.50} \\

stack\_blocks\_two
& 56.00 & \textbf{73.00} & 62.00 & \textbf{76.00} & 59.00 & \textbf{74.50} \\

stack\_bowls\_three
& 50.00 & \textbf{55.00} & 54.00 & \textbf{55.00} & 52.00 & \textbf{55.00} \\

stack\_bowls\_two
& \textbf{85.00} & 78.00 & \textbf{83.00} & 72.00 & \textbf{84.00} & 75.00 \\

stamp\_seal
& \textbf{55.00} & 51.00 & 51.00 & \textbf{52.00} & \textbf{53.00} & 51.50 \\

turn\_switch
& \textbf{54.00} & 46.00 & \textbf{59.00} & 44.00 & \textbf{56.50} & 45.00 \\

\midrule
\textbf{Average}
& 66.42 & \textbf{69.66} & 67.42 & \textbf{67.54} & 66.92 & \textbf{68.60} \\
\bottomrule
\end{tabular}
}
\end{table*}
\clearpage

\begin{table*}[t]
\centering
\caption{
Ablation of the AdaIE equalization strength $\alpha$ on LeWM (seed 42).
We report planning success rates (\%) at each training epoch. \emph{None} denotes the baseline without AdaIE.
Each checkpoint is evaluated over 50 episodes using the same evaluation seed.
}
\small
\setlength{\tabcolsep}{3.2pt}
\resizebox{\textwidth}{!}{
\begin{tabular}{ll*{10}{c}}
\toprule
\textbf{Task}
& \textbf{$\alpha$}
& \textbf{Ep. 1}
& \textbf{Ep. 2}
& \textbf{Ep. 3}
& \textbf{Ep. 4}
& \textbf{Ep. 5}
& \textbf{Ep. 6}
& \textbf{Ep. 7}
& \textbf{Ep. 8}
& \textbf{Ep. 9}
& \textbf{Ep. 10} \\
\midrule

\multirow{8}{*}{Two Room}
& None & 94 & 90 & 88 & 90 & 88 & 88 & 90 & 88 & 90 & 92 \\
& 0.05 & 94 & 94 & 94 & 90 & 94 & 94 & 92 & 94 & 92 & 90 \\
& 0.10 & 92 & 90 & 90 & 90 & 88 & 90 & 92 & 92 & 92 & 92 \\
& 0.15 & 94 & 94 & 92 & 88 & 92 & 90 & 90 & 92 & 94 & 92 \\
& 0.25 & 92 & 92 & 90 & 88 & 92 & 94 & 94 & 94 & 94 & 94 \\
& 0.35 & 92 & 90 & 88 & 88 & 92 & 94 & 94 & 92 & 96 & 94 \\
& 0.40 & 96 & 92 & 88 & 90 & 92 & 88 & 92 & 92 & 94 & 94 \\
& 0.50 & 96 & 92 & 92 & 92 & 94 & 94 & 94 & 92 & 94 & 92 \\
\midrule

\multirow{8}{*}{Reacher}
& None & 66 & 64 & 78 & 76 & 76 & 80 & 80 & 80 & 86 & 80 \\
& 0.05 & 62 & 70 & 78 & 76 & 66 & 80 & 82 & 86 & 76 & 74 \\
& 0.10 & 74 & 78 & 74 & 82 & 70 & 80 & 76 & 84 & 72 & 78 \\
& 0.15 & 72 & 66 & 64 & 76 & 68 & 86 & 76 & 80 & 84 & 82 \\
& 0.25 & 76 & 68 & 80 & 74 & 70 & 78 & 76 & 76 & 76 & 84 \\
& 0.35 & 72 & 70 & 66 & 78 & 72 & 82 & 84 & 78 & 80 & 74 \\
& 0.40 & 66 & 76 & 78 & 86 & 70 & 80 & 78 & 78 & 84 & 84 \\
& 0.50 & 58 & 78 & 82 & 76 & 74 & 86 & 80 & 78 & 84 & 84 \\
\midrule

\multirow{8}{*}{Push-T}
& None & 66 & 82 & 88 & 90 & 88 & 88 & 90 & 92 & 92 & 88 \\
& 0.05 & 64 & 86 & 88 & 86 & 92 & 90 & 92 & 92 & 96 & 96 \\
& 0.10 & 68 & 74 & 90 & 88 & 90 & 96 & 92 & 96 & 94 & 96 \\
& 0.15 & 80 & 86 & 86 & 94 & 94 & 96 & 100 & 94 & 98 & 94 \\
& 0.25 & 72 & 86 & 90 & 92 & 96 & 94 & 94 & 98 & 98 & 94 \\
& 0.35 & 66 & 84 & 84 & 92 & 98 & 94 & 96 & 94 & 94 & 92 \\
& 0.40 & 64 & 84 & 86 & 94 & 90 & 94 & 90 & 98 & 94 & 96 \\
& 0.50 & 62 & 76 & 84 & 86 & 94 & 84 & 90 & 98 & 90 & 94 \\
\midrule

\multirow{8}{*}{OGBench-Cube}
& None & 70 & 72 & 72 & 76 & 72 & 76 & 70 & 70 & 70 & 74 \\
& 0.05 & 78 & 76 & 72 & 76 & 80 & 80 & 78 & 76 & 76 & 80 \\
& 0.10 & 76 & 78 & 78 & 76 & 78 & 80 & 78 & 82 & 78 & 78 \\
& 0.15 & 72 & 68 & 72 & 74 & 72 & 76 & 74 & 72 & 72 & 72 \\
& 0.25 & 76 & 80 & 74 & 78 & 76 & 80 & 82 & 78 & 80 & 80 \\
& 0.35 & 80 & 78 & 82 & 80 & 82 & 84 & 80 & 74 & 78 & 78 \\
& 0.40 & 78 & 76 & 82 & 74 & 72 & 78 & 78 & 76 & 78 & 78 \\
& 0.50 & 80 & 76 & 78 & 78 & 80 & 78 & 80 & 76 & 78 & 78 \\
\bottomrule
\end{tabular}
}
\label{tab:lewm_alpha_full}
\end{table*}

\clearpage
\subsection{Co-adaption on other models}
\label{app:co-adaption}
\subsubsection{Diffusion Policy}
\begin{figure}[H]
    \centering
    \includegraphics[width=\linewidth]{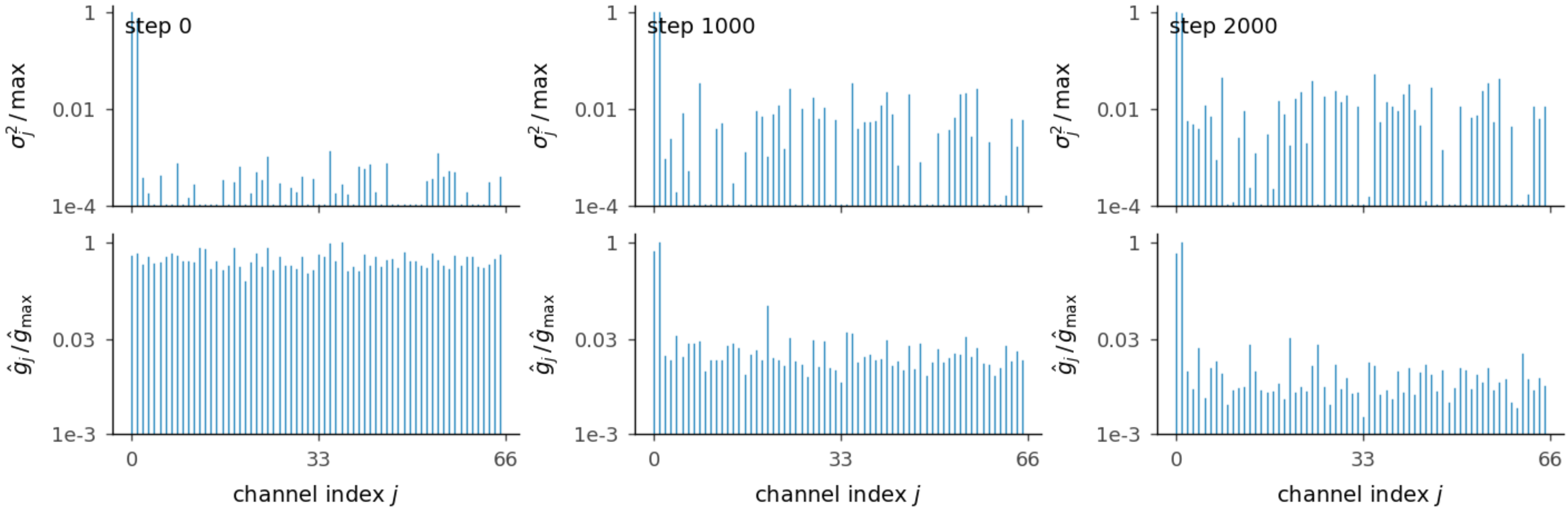}
    \caption{The co-adaption of DiffusionPolicy-C. The Policy is trained on Push-T.}
    \label{fig:dp_c_g_vs_s}
\end{figure}

\begin{figure}[H]
    \centering
    \includegraphics[width=\linewidth]{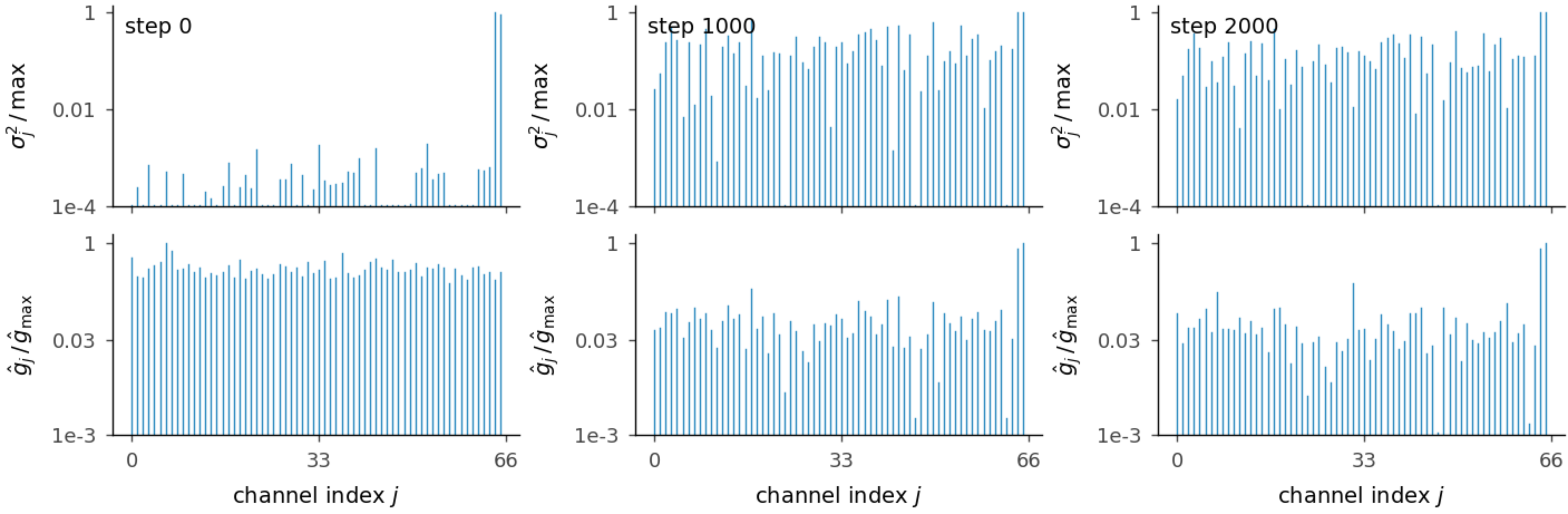}
    \caption{The co-adaption of DiffusionPolicy-T.The Policy is trained on Push-T.}
    \label{fig:dp_t_g_vs_s}
\end{figure}
\subsubsection{DINO-WM}
\begin{figure}[H]
    \centering
    \includegraphics[width=\linewidth]{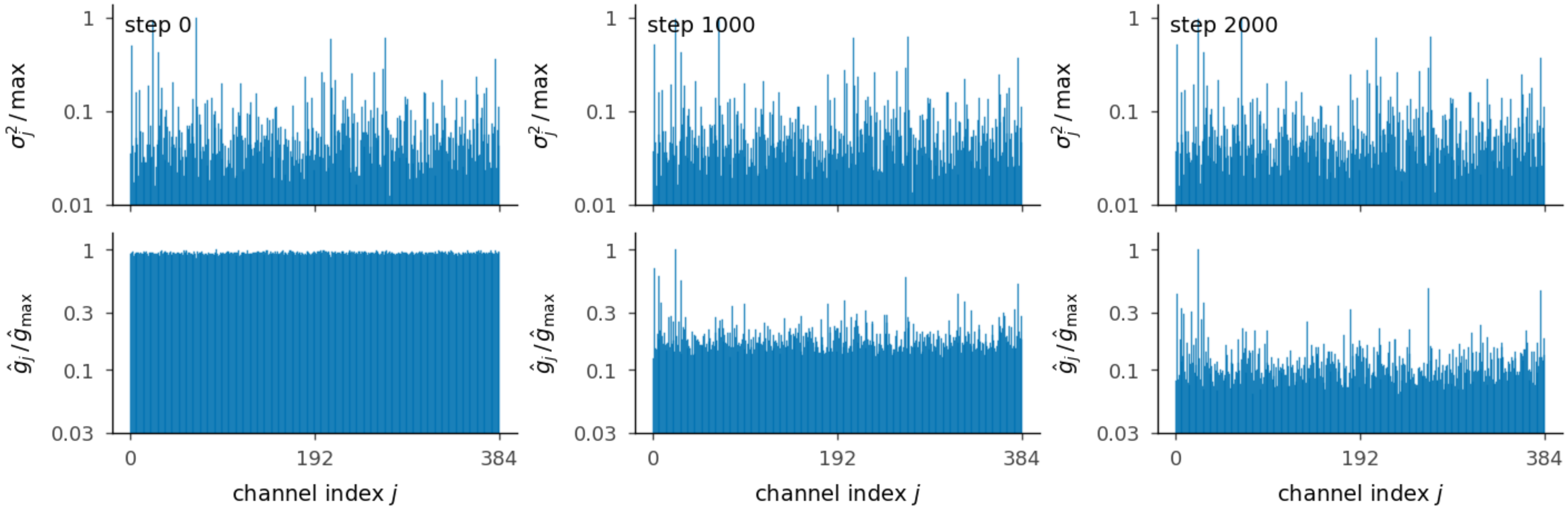}
    \caption{The co-adaption of DINO-WM.The model is trained on Two Room.}
    \label{fig:dinowm_g_vs_s}
\end{figure}

\end{document}

%% file: math_commands.tex
\usepackage{amsmath,amsfonts,bm}

\def\eqref#1{equation~\ref{#1}}

\def\1{\bm{1}}

\DeclareMathAlphabet{\mathsfit}{\encodingdefault}{\sfdefault}{m}{sl}
\SetMathAlphabet{\mathsfit}{bold}{\encodingdefault}{\sfdefault}{bx}{n}

